# MLOps with enhanced performance control and observability

Indradumna Banerjee
Principal Data Scientist
Indradumna.banerjee@oracle.com

Dinesh Ghanta
Principal Data Scientist
dinesh.ghanta@oracle.com

Girish Nautiyal
Senior Data Scientist
girish.nautiyal@oracle.com

Pradeep Sanchana
Principal Software Engineer
pradeep.sanchana@oracle.com

Prateek Katageri
Principal Machine Learning Engineer
prateek.katageri@oracle.com

Atin Modi
Applications Engineer
atin.modi@oracle.com

## ABSTRACT

The explosion of data and its ever increasing complexity in the last few years, has made MLOps systems more prone to failure, and new tools need to be embedded in such systems to avoid such failure. In this demo, we will introduce crucial tools in the observability module of a MLOps system that target difficult issues like data drfit and model version control for optimum model selection. We believe integrating these features in our MLOps pipeline would go a long way in building a robust system immune to early stage ML system failures.

## CCS CONCEPTS

• MLOps Monitoring and control

## Introduction

Machine Learning has been widely adopted by the industry in the last few years, leading to scalability, innovativeness, efficiency, and sustainability of multiple businesses. However, the complex evolving nature of data means there are new challenges to solve, with a majority of ML proof-of-concept models failing when taken to production systems. One of the main reasons for these failures is lesser emphasis of coordinating the complex ML system components and infrastructure, in an automated way across many industrial applications. [1,2]

So, our efforts in building automatic, productive, and operationalized ML systems in production, is about tackling three main essential challenges. These challenges are preprocessing large amounts of data, setting up tracking and versioning for experiments and model training runs, and setting up observability in the deployment pipelines for models running in production. The need for solution to these challenges led to the evolution of MLOps, which is built from DevOps, Data Engineering, and Machine Learning. As an approach and tool for ML lifecycle management, there has been a few reports in the past, [3,4] where end to end MLOps systems has been demonstrated. In this demo, we will present an improved end to end MLOps system, where novel tools developed for addressing issues such as data drift, concept drift and model version control, have been integrated into the system. We will also be discussing these issues in detail in our demo and explain why they are crucial problems worth solving in any MLOps system.

## System Design

An MLOps system needs to be designed with an understanding that the system is the actual product, not the ML model and it components. [5,6,7] With this perspective, we have listed down the main components of an MLOps system in Fig.1, that includes an Orchestrator, User Interface (UI), Data Source, ML components, Data Version Control (DVC), Metric collectors, database (DB). We also introduce a new module on observability and enable notifications that work as an alert monitoring system.

The MLOps Orchestrator, helps in streamlining and enforcing a robust architecture, and works towards ensuring high system availability. Having a well-designed ML user interface, that captures all aspects of the features developed in code is extremely crucial, as it helps in better explainability of what the system is doing and explain its unique features. Machine Learning Components in the system includes several features such as data preprocessing, model training and scoring, model inference, and these components are the core of any standard MLOps platform. Data Version Control (DVC) is about managing data and ML experiments, and it leverages existing versioning and engineering toolset such as Git and CI/CD. While Git is used for storing and version code, DVC on similar lines stores the data and model files.

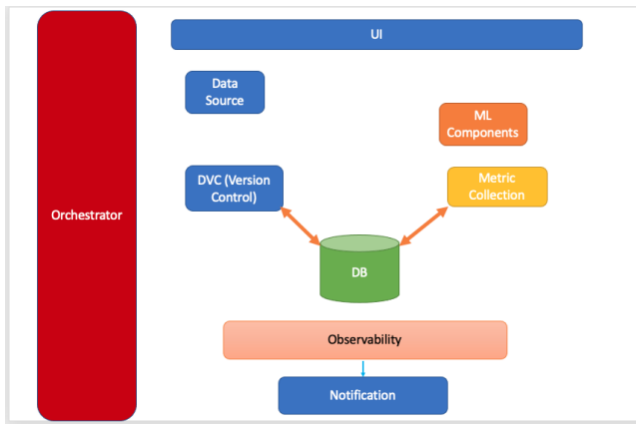

**Figure 1: Components of a designed MLOps system, with enhanced and novel observability module that includes crucial components such as data and concept drift monitoring, and model experimental tracking.**

For ML systems in production, we often rely on cloud resources, components, and services, that provide log and metrics. [8,9] The monitoring metrics provide an efficient mechanism to perform root cause analysis in troubleshooting production and development issues. Finally, we would be demonstrating an enhanced Observability module and we introduce two key features in this module: Model Experimentation System (MES) and a data validation module.

*Model Experimentation System (MES):* We have added a unique model experimentation system, that will allow users to pick out the best versions of a particular model, and analyse which changes in the parameters made the performance better or worse. This feature is intended to help users recreate the results of a particular model version because a model lineage will be stored for each model version. The model lineage will also aid users in understanding how and why the different model versions were made. The MES will be useful during A/B/n testing, multivariate testing and statistical analysis of test results to determine which model version performs the best. It also offers transparency on the models itself and more details about its feature inputs and performance outputs.

*Data Validation module:* Machine Learning models over time degenerate in their performance and accuracy, primarily due to a shift in the patterns of the data the model uses for it predictions. This shift in data patterns is known as Data Drift, which is important to identify at an early stage. Identification of Data Drift is the main feature of Data Validation module, that also performs basic schema level checks and high level feature wise checks. Drift detection module consists of five main sub-modules: services, data format identifier, summary generator, benchmark and Output Interpreter. All the sub-module work together towards a bigger goal of maintaining the model performance over time, and generating real-time alerts. Services sub module uses the configurable parameters in all other sub-modules, and updates the configurable parameters. Data format identifier uses the dataset and user config json file, and generates the categorisation of features based on user provided input thresholds. Summary generator sub-module generates summary for multiple types of features that includes for example number of records in the schema, and features related to numerical cardinality features amongst others. The benchmark sub-module compares the various metrics generated in the summary module, and monitors the relative change from baseline value to the overall percentage data drift accepted. Finally, the output interpreter sub-module generates alerts and other insights for the benchmark sub-module, in a UI presentable format.

## Conclusion

With an increased impetus on ML innovation coupled with data availability and analytical capabilities, the need for machine learning products working as a time tested system is extreme. The translation of this need into effect, however, needs a lot of integrated efforts for firstly progressing ML models into deployment and production, and then, more importantly maintaining them at a constant performance level in production. Machine Learning Operations (MLOps) is a promising solution on this front, and in this demo, we address some key challenges towards building a robust MLOps system. We demonstrate our entire system, and its key unique features, that should promote better monitoring and performance control of MLOps systems in production.

The demo aims to build a common understanding of an MLOps system and its associated concepts, and hopefully should assist businesses and individuals in setting up non-degenerating MLOps systems, by using the novel features related to robust MLOps systems that we introduce.